\pdfoutput=1

\documentclass[11pt]{article}

\usepackage{ACL2023}

\usepackage{times}
\usepackage{latexsym}
\usepackage{graphicx}
\usepackage[T1]{fontenc}

\usepackage{booktabs}
\usepackage[utf8]{inputenc}

\usepackage{microtype}

\usepackage{inconsolata}

\usepackage{multirow}
\usepackage{tablefootnote}
\usepackage{float}

\usepackage{pifont}
\usepackage{color, colortbl}

\definecolor{tablegray}{rgb}{0.9,0.9,0.9}
\definecolor{darkgreen}{rgb}{0,0.5,0}
\definecolor{lightblue}{rgb}{0.7,0.9,1}
\definecolor{lblue}{rgb}{0.25,0.5,1}

\usepackage[normalem]{ulem}

\def\ODdel#1{\bgroup\markoverwith{\textcolor{red}{\rule[0.4ex]{2pt}{3pt}}}\ULon{#1}}
\def\VDdel#1{\bgroup\markoverwith{\textcolor{green!90!red!55}{\rule[0.4ex]{2pt}{3pt}}}\ULon{#1}}
\newcommand{\cmark}{\ding{51}}%
\newcommand{\xmark}{\ding{55}}%

\title{Are Large Language Models All You Need for Task-Oriented Dialogue?}

\author{Vojtěch Hudeček \and Ondřej Dušek \\
Charles University, Faculty of Mathematics and Physics\\Malostranské náměstí 25, 118 00 Prague, Czechia\\
\texttt{hudecek@ufal.mff.cuni.cz}, \texttt{odusek@ufal.mff.cuni.cz}
}

\begin{document}
\maketitle
\begin{abstract}
Instruction-finetuned large language models (LLMs) gained a huge popularity recently, thanks to their ability to interact with users through conversation. In this work, we aim to evaluate their ability to complete multi-turn tasks and interact with external databases in the context of established task-oriented dialogue benchmarks.
We show that in explicit belief state tracking, LLMs underperform compared to specialized task-specific models.
Nevertheless, they show some ability to guide the dialogue to a successful ending through their generated responses if they are provided with correct slot values.
Furthermore, this ability improves with few-shot in-domain examples.
\end{abstract}

\section{Introduction}
\label{sec:intro}
Large Language Models (LLMs) have transformed the NLP field,
showing outstanding performance across many NLP benchmarks such as Winograd Challenge \cite{levesque2012winograd} or GLUE \cite{wang2018glue}.
Recently, instruction finetuning of LLMs proved to be able to align the model outputs with human preferences \cite{ouyang2022training,supernaturalinstructions} and improved the LLMs' communication capabilities substantially.
State-of-the-art LLMs are not only good at understanding user needs but also capable of providing relevant answers.
Consequently, we see many chatbot applications both inside and outside academia (ChatGPT\footnote{\url{https://openai.com/blog/chatgpt}}, Claude\footnote{\url{https://www.anthropic.com/index/introducing-claude}}, Sparrow\footnote{\url{https://www.deepmind.com/blog/building-safer-dialogue-agents}}) which build upon the raw power of instruction-finetuned LLMs.

\begin{figure}[t!]
    \centering
    \includegraphics[width=0.4\textwidth]{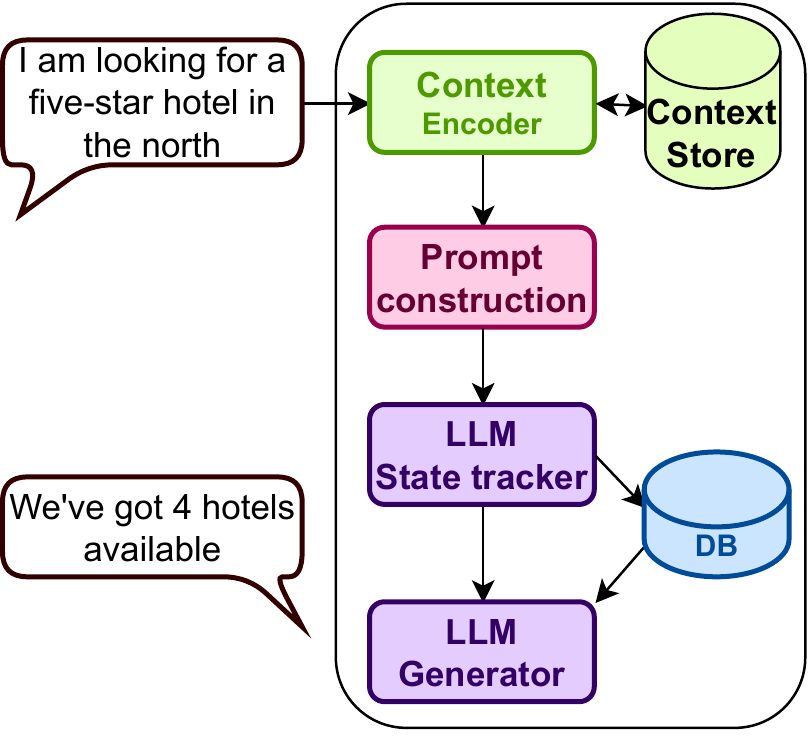}
    \caption{A high-level overview of our proposed pipeline. The user input is used to retrieve relevant few-shot examples (if available). Then, an initial prompt is constructed and an LLM is asked to provide the current dialogue state. Based on that, we retrieve database results and construct another prompt. Finally, we ask the LLM to provide the response.}
    \label{fig:overview_high_level}
\end{figure}
\begin{figure*}[t!]
    \centering
    \includegraphics[width=\textwidth]{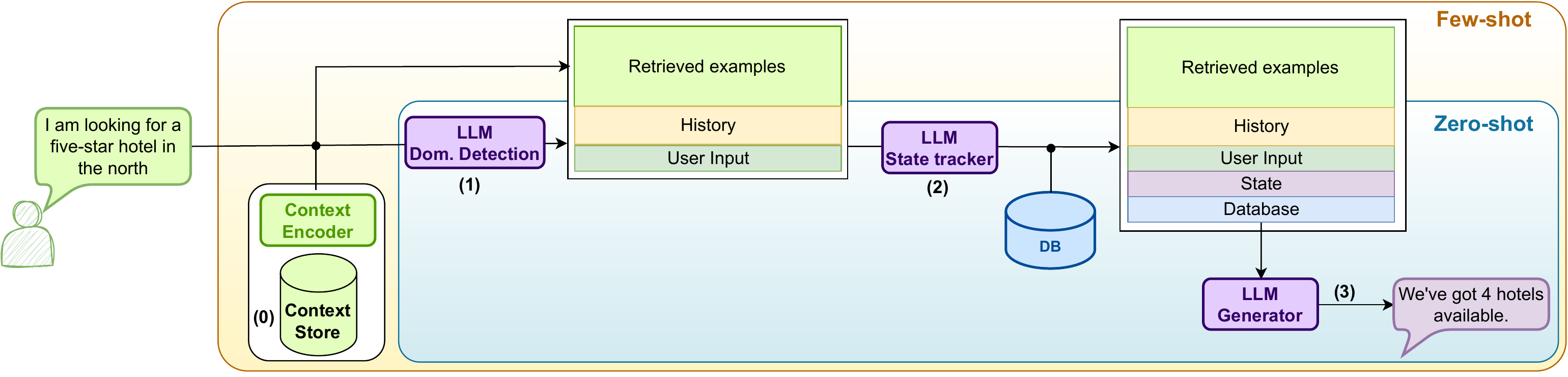}
    \caption{A detailed description of our proposed pipeline. (0) As a preprocessing step, we encode a subset of the training set that will be used to retrieve few-shot examples.
    Given the user input, we: (1) Detect the domain, retrieve relevant examples (in the few-shot setting) and construct an initial prompt. (2) Infer the belief state using LLM. Based on that, we retrieve database information and construct another prompt that includes both the state and database results. (3) We ask the LLM to provide a final response.}
    \label{fig:overview_low_level}
\end{figure*}

Given the millions of daily interactions with these chatbots, it appears that the models are able to handle users' needs to their satisfaction, at least to some extent.
However, these chatbots are tuned using unstructured open-domain conversations.
The aim of this paper is to evaluate these systems
for more specific applications, where the system has to follow a predetermined structure and handle external sources of information, such as APIs or databases.
We raise the question to what extent LLMs are capable of handling these applications off-the-shelf, i.e.\ without finetuning.
We thus choose to evaluate LLM performance in the task-oriented dialogue (TOD) setting,
as it requires precise information handling for communicating with external APIs.
Moreover, TOD systems output in-domain information which has predetermined structure and lends itself well to evaluation, thanks to pre-existing annotated data sets.
We avoid any finetuning techniques and focus on zero-shot or few-shot settings using in-context learning, as this approach has lower hardware requirements and barrier of entry and better flexibility or even performance in certain tasks \cite{su2022selective}. 

Therefore, we introduce an LLM-based TOD conversation pipeline (see Figure \ref{fig:overview_high_level}) and evaluate its performance with respect to commonly used task-oriented metrics such as Joint Goal Accuracy, Slot F1, and Dialogue Success \cite{rastogi_multi-task_2018,budzianowski_multiwoz_2018}.
Our pipeline resembles other approaches based on LMs \cite{peng-etal-2021-soloist,yang2021ubar}, using state tracking and response generation as two main, separate steps, while keeping the role of a dialogue policy implicit.
However, instead of finetuning LMs, it intentionally relies almost exclusively on the usage of pretrained LLMs as-is, so we can test their out-of-the-box capabilities.
The dialogue context and domain description are introduced to the model only by including them in the input prompt.
In the zero-shot setting, the model receives a domain description only; in the few-shot setting, it additionally uses a few retrieved examples (see Section~\ref{sec:method} for details).

In our experiments, we find that LLMs are not very good at state tracking and their performance falls behind the state-of-the-art. 
However, if provided with correct belief states, some of them yield interesting response generation performance, comparable to earlier finetuned state-of-the-art models. 
To our knowledge, our zero-shot experiments establish a state-of-the-art result in unsupervised TOD modeling on the MultiWOZ and Schema-guided datasets \cite{budzianowski_multiwoz_2018,rastogi2020towards}.
{While there may be room for improvement through prompt engineering, our results aim to show the out-of-the-box LLM capabilities.
We plan to release our experimental code on GitHub.\footnote{\url{https://github.com/vojtsek/to-llm-bot}}

\section{Related Work}
\label{sec:related}

\paragraph{Large Language Models}
The Transformer architecture \cite{vaswani2017attention} enabled the training of large and capable language models.
The research on their few-shot and zero-shot abilities dates back to the GPT-2 and GPT-3 models \cite{radford2019language,brown2020language}, which are scaled versions of the Transformer decoder.
Many followed this path of training large Transformer decoders \cite{zhang2022opt,black2022gpt}, yielding models of up to hundreds of billions parameters in size \cite{zhao_survey_2023}.
Other models leverage the whole original (encoder-decoder) Transformer architecture \cite{2020t5,lewis_bart:_2020}.
Recent research focuses on improving the training of moderate-sized architectures to broaden access to highly capable LLMs \cite{touvron2023llama}.

\paragraph{Instruction Tuning}
The idea of using reinforcement learning techniques to align model-based agents better with users' intents was pioneered in game agent development \cite{christiano2017deep} and later explored for training language models \cite{ziegler2019fine,ouyang2022training}.
Although these techniques proved to be quite effective, the process is still very demanding in terms of collecting feedback from users.
Consequently, several datasets were proposed \cite{supernaturalinstructions,iyer2022opt,black2022gpt} that collected millions of instructions-based tasks in natural language and can be applied to finetune LLMs using reinforcement learning.

\paragraph{LM-based TOD modeling}
Task-oriented dialogue modeling with pretrained LMs was introduced by \citet{zhang2019dialogpt} and \citet{peng-etal-2021-soloist}, who followed text-based state encoding and two-stage generation proposed by \citet{lei2018sequicity}:
An LM is first used to decode a structured belief state, represented as text.
The belief state is then used to retrieve database information and the LM is called once more to generate a response, conditioned on the belief state and retrieved information.
Several improvements to the basic setup were proposed, such as contrastive state training \cite{kulhanek-etal-2021-augpt} or using belief state differences \cite{lin-etal-2020-mintl}.
Others proposed a combination of generative models with retrieval-based approaches \cite{pandey-etal-2018-exemplar,cai-etal-2019-retrieval,nekvinda-dusek-2022-aargh}.
All described works finetune LMs on in-domain data, which is in contrast with the pure in-context learning approach that we apply.

\paragraph{Few-shot dialogue modeling}
One of the first neural models focusing on learning dialogue from a few in-domain examples was the Hybrid Code Networks  \cite{williams-etal-2017-hybrid}, a trainable system based on recurrent neural networks, with partially handcrafted components.
Another approach was proposed by \citet{zhao-eskenazi-2018-zero}, who used latent action representations to enable the transfer of domain knowledge.
Latent actions were also used by \citet{huang2020mala} and \citet{shalyminov-etal-2019-data}.
More recent approaches leverage the capabilities of pretrained Transformer LMs \cite{shalyminov_fast_2020}.
\citet{hu-etal-2022-context} used LLMs and in-context learning to perform belief state tracking, formulating the task as an SQL query generation.
Unlike our work, they did not use instruction-tuned models and omitted database retrieval and response generation.

\section{Method}
\label{sec:method}

We introduce our method step-by-step.
An overall description of the proposed pipeline is shown in Figure~\ref{fig:overview_low_level}.
The system consists of a pretrained LLM and an (optional) context store in a vector database.
Three LLM calls are performed in each dialogue turn, with specific prompts (see Section~\ref{sec:prompts}).
First, the LLM performs domain detection and state tracking (Section~\ref{sec:state-tracking}). The updated belief state informs a database query, whose results are used in the subsequent LLM-based response generation step (Section~\ref{sec:response-generation}).
In the few-shot setting, the context store is used to store a limited number of examples from the training set, which are retrieved based on similarity with the conversation context and included in LLM prompts (see Section~\ref{sec:context-store}).


\begin{table}[t!]
    \centering\small
    \begin{tabular}{rl}
      \toprule
      \textbf{Prompt} & {\color{cyan!80!yellow!80!black!100}Definition: Capture values from a conversation } \\
      & {\color{cyan!80!yellow!80!black!100} about hotels. } \\
      & {\color{cyan!80!yellow!80!black!100} Capture pair "entity:value" separated by colon } \\
      & {\color{cyan!80!yellow!80!black!100} and no spaces in between.} \\
      & {\color{cyan!80!yellow!80!black!100} Separate the "entity:value" pairs by hyphens } \\
      & {\color{cyan!80!yellow!80!black!100} Values that should be captured are: } \\
      & {\color{green!100!yellow!70!black!100!} - "pricerange": the price of the hotel} \\
      &  ... \\
      & {\color{red!100!yellow!70!black!100!}[history] } \\
      &  {\color{orange!50!yellow!90!black!100!}Customer: "I want a cheap place to stay." }\\
      \midrule
      \textbf{Output:} & pricerange:"cheap"\\
      \bottomrule
  \end{tabular}
  \caption{A simplified example of a zero-shot version of the prompt used for state update prediction.
  It contains {\color{cyan!80!yellow!80!black!100} task definition},  {\color{green!100!yellow!70!black!100!}domain description}, {\color{red!100!yellow!70!black!100!} dialogue history} and {\color{orange!50!yellow!90!black!100!} user utterance}.
  For the exact prompts see Appendix.}
  \label{tab:state_tracking}
\end{table}

\subsection{Prompt construction}
\label{sec:prompts}
We aim to compare the raw capabilities of the selected LLMs, therefore we do not focus on prompt engineering techniques and choose universal prompts used for all LLMs in this work (cf.~Section~\ref{sec:limit}).
We choose simple, plain language statements as prompts, with no specific vocabulary, based only on a few preliminary tests.
We define a single \textbf{domain detection prompt} for all examples, plus a pair of prompts for each domain in the given dataset: a \textbf{state tracking prompt} (see Table~\ref{tab:state_tracking}) and a \textbf{response prompt}.

The domain detection prompt includes a task description and two static examples of domain detection.
In addition to general instructions, each state tracking prompt contains a domain description, a list of relevant slots, the dialogue history, and the current user utterance.
The response prompts do not contain the per-domain slot list, but they include the current belief state and database results instead.
In the few-shot setting, each tracking and response prompt additionally contains positive and negative examples retrieved from the context store (see Section~\ref{sec:context-store}).
Prompt examples are shown in Tables~\ref{app:zero-shot-state} and~\ref{app:zero-shot-response} in the Appendix.

\subsection{Domain Detection and State Tracking}
\label{sec:state-tracking}

We prompt the LM twice at each turn during state tracking: first, to detect the active domain, then to output slot values that changed or appeared in the current turn. We then use the outputs to update the accumulated global belief state.

The two prompting steps are used since 
we need the models to operate in a multi-domain setting, i.e., handle conversations spanning multiple domains.
Therefore, we need to be able to detect the currently active domain.
We achieve this by first prompting the LLM with a domain detection prompt (using a single prompt for all examples).

Once we obtain the active domain prediction, we can include manually designed domain descriptions in a second prompt that handles belief state prediction.
An example of a prompt used for state tracking is provided in Table~\ref{tab:state_tracking}.
For the few-shot variants, 
we retrieve few-shot examples from the context store, limited to the active domain.\footnote{For this purpose, each conversation snippet contained in the context store comes from a single-domain conversation.}

Our preliminary experiments showed that LLMs struggle to output all active slot values at every turn consistently.
Therefore, we model only state updates, following the MinTL approach \cite{lin-etal-2020-mintl}.
Here, the model only generates the slot-value pairs that have changed in current turn.
The global belief state is then accumulated using these turn-level updates.
To obtain machine-readable outputs useful for database queries or API calls,
we specify in the prompt that the model should provide JSON outputs, and any provided few-shot examples are formatted accordingly.

\begin{table*}[t]
    \centering\small
    \begin{tabular}{l|c|c|ccc>{\hspace{-2mm}}c|ccc>{\hspace{-2mm}}c}
      \toprule
      model & few & oracle & \multicolumn{4}{c|}{\textbf{Schema Guided Dialogues}} & \multicolumn{4}{c}{\textbf{MultiWOZ 2.2}} \\
      & shot & BS & BLEU & JGA & Slot-F1 & Success & BLEU & JGA & Slot-F1 & Success \\
      \midrule
      Supervised SotA & \textcolor{red}{\xmark} & \textcolor{red}{\xmark} & 29.90$^\ast$ & 0.30$^\dagger$ & 0.60$^\ast$ & -- & 19.90$^\clubsuit$ & 0.60$^\diamondsuit$ & -- & 0.82$^\heartsuit$ \\
      \midrule
      \rowcolor{tablegray}
      \emph{Alpaca-LoRA-7B-zs-gbs} & \textcolor{red}{\xmark} & \textcolor{red}{\xmark} & 2.79 & 0.02 & 0.01 & 0.11 & 1.61 & 0.06 & 0.07 & 0.04 \\
      \rowcolor{tablegray}
      \emph{Tk-Instruct-11B-zs-gbs} & \textcolor{red}{\xmark} & \textcolor{red}{\xmark} & 4.16 & 0.05 & 0.03 & 0.10 & 2.48 & 0.04 & 0.04 & 0.04 \\
      \rowcolor{tablegray}
      \emph{GPT-NeoXT-20B-zs-gbs} & \textcolor{red}{\xmark} & \textcolor{red}{\xmark} & 0.45 & 0.01 & 0.01 & 0.17 & 0.52 & 0.03 & 0.02 & 0.04 \\
      \rowcolor{tablegray}
      \emph{OPT-IML-30B-zs-gbs} & \textcolor{red}{\xmark} & \textcolor{red}{\xmark} & 1.63 & 0.01 & 0.01 & 0.17 & 0.56 & 0.02 & 0.04 & 0.03 \\
      \rowcolor{tablegray}
      \emph{ChatGPT-zs-gbs} & \textcolor{red}{\xmark} & \textcolor{red}{\xmark} & -- & -- & -- & -- & 4.17 & 0.13 & 0.40 & 0.31 \\ 

      \emph{Alpaca-LoRA-7B-zs-obs} & \textcolor{red}{\xmark} & \textcolor{green}{\cmark} & 2.76 & -- & -- & 0.23 & 1.73 & -- & -- & 0.08 \\
      \emph{Tk-Instruct-11B-zs-obs} & \textcolor{red}{\xmark} & \textcolor{green}{\cmark} & 5.21 & -- & -- & 0.24 & 2.66 & -- & -- & 0.18 \\
      \emph{GPT-NeoXT-20B-zs-obs} & \textcolor{red}{\xmark} & \textcolor{green}{\cmark} & 0.83 & -- & -- & 0.22 & 0.60 & -- & -- & 0.06 \\
      \emph{OPT-IML-30B-zs-obs} & \textcolor{red}{\xmark} & \textcolor{green}{\cmark} & 1.94 & -- & -- & 0.22 & 0.54 & -- & -- & 0.06 \\
      \emph{ChatGPT-zs-obs} & \textcolor{red}{\xmark} & \textcolor{green}{\cmark} & -- & -- & -- & -- & 3.76 & -- & -- & 0.47 \\ \midrule

      \rowcolor{tablegray}
      \emph{Alpaca-LoRA-7B-fs-gbs} & \textcolor{green}{\cmark} & \textcolor{red}{\xmark} & 6.32 & 0.04 & 0.01 & 0.09 & 5.53 & 0.06 & 0.08 & 0.06\\
      \rowcolor{tablegray}
      \emph{Tk-Instruct-11B-fs-gbs} & \textcolor{green}{\cmark} & \textcolor{red}{\xmark} & 6.66 & 0.06 & 0.05 & 0.10 & 6.56 & 0.16 & 0.33 & 0.19 \\
      \rowcolor{tablegray}
      \emph{GPT-NeoXT-20B-fs-gbs} & \textcolor{green}{\cmark} & \textcolor{red}{\xmark} & 1.62 & 0.04 & 0.02 & 0.09 & 2.73 & 0.05 & 0.04 & 0.05 \\
      \rowcolor{tablegray}
      \emph{OPT-IML-30B-fs-gbs} & \textcolor{green}{\cmark} & \textcolor{red}{\xmark} & 0.82 & \textbf{0.06} & \textbf{0.07} & 0.08 & 4.40 & 0.03 & 0.03 & 0.04 \\
      \rowcolor{tablegray}
      \emph{ChatGPT-fs-gbs} & \textcolor{green}{\cmark} & \textcolor{red}{\xmark} & -- & -- & -- & -- & 6.77 & \textbf{0.27} & \textbf{0.51} & 0.44 \\

      \emph{Alpaca-LoRA-7B-fs-obs} & \textcolor{green}{\cmark} & \textcolor{green}{\cmark} & 6.99 & -- & -- & \textbf{0.25} & 5.96 & -- & -- & 0.41 \\
      \emph{Tk-Instruct-11B-fs-obs} & \textcolor{green}{\cmark} & \textcolor{green}{\cmark} & \textbf{8.56} & -- & -- & \textbf{0.25} & \textbf{6.91} & -- & -- & 0.46 \\
      \emph{GPT-NeoXT-20B-fs-obs} & \textcolor{green}{\cmark} & \textcolor{green}{\cmark} & 1.97 & -- & -- & 0.24 & 2.92 & -- & -- & 0.28 \\
      \emph{OPT-IML-30B-fs-obs} & \textcolor{green}{\cmark} & \textcolor{green}{\cmark} & 0.56 & -- & -- & 0.22 & 5.40 & -- & -- & 0.28 \\
      \emph{ChatGPT-fs-obs} & \textcolor{green}{\cmark} & \textcolor{green}{\cmark} & -- & -- & -- & -- & 6.84 & -- & -- & \textbf{0.68} \\

    \bottomrule
  \end{tabular}
  \caption{
  Evaluation of the chosen LLMs with respect to widely used TOD measures. For each model, we provide multiple variants. We use either zero-shot or few-shot prompts (\emph{-zs-} vs. \emph{-fs-}) and either generated or oracle belief state (\emph{-gbs} vs. \emph{-obs}).
  The few-shot variants use 10 examples per domain in the context storage ($\sim$0.6\% of the training set in case of MultiWOZ), two of which are selected for the prompts.
  To reduce cost, we only evaluate the paid ChatGPT model on MultiWOZ.
  We also provide supervised state-of-the-art results to put the numbers in context: $^\ast$\citet{zhu2022convlab3}, $^\dagger$\citet{feng-etal-2021-sequence}, $^\clubsuit$\citet{sun2022mars}, $^\diamondsuit$\citet{huangrobustness}, $^\heartsuit$\citet{feng2023fantastic}. }
  \label{tab:res_overall}
\end{table*}

\subsection{Response Generation}
\label{sec:response-generation}

The current belief state is used to query the database for entries matching all user-specified slots in the active domain. Given the belief state and database results, the response generation is straightforward.
The prompt for the LLM includes dialogue history, user utterance, belief state and database results (and retrieved examples in the few-shot setting) and requests the model to provide a fitting system response.
We generate delexicalized responses \cite{wen-etal-2015-stochastic}, i.e., we replace slot values by placeholders, following prior work in end-to-end TOD modeling.
In addition to simplifying the task for the model, delexicalized outputs allow us to evaluate the success rate and compare to previous works.
The prompt specifies that the model should provide entity values as delexicalized placeholders, and any few-shot examples are constructed accordingly.

\subsection{Context Storage}
\label{sec:context-store}

It has been shown that enriching prompts with specific examples boosts LM performance \cite{madotto2020language,brown2020language}.
To apply this knowledge efficiently in our pipeline, we introduce a storage that contains encoded dialogue contexts.
This context storage is optional and is only required for the few-shot prompting variant.
We use dialogue context taken from a fixed-length history window as the key to be encoded in the vector database.
More details can be found in Section~\ref{subsec:exp-details}.
Once the relevant examples are retrieved, we include them in the prompt to guide the model better.
Some of the LLMs rely on negative (counter-) examples as well \cite{supernaturalinstructions}. Therefore, we follow \citet{peng-etal-2021-soloist}'s consistency classification task approach to produce negative examples: We take some of the retrieved belief state examples, corrupt them by replacing some of the correct slot values with random values, and present them as negative in the prompt.

\section{Experimental Setup}
\label{sec:experiments}

To obtain a broad overview of the current LLMs' capabilities, we compare several models, spanning different numbers of trainable parameters and different training methods. 
We also experiment with four variants of the base setup, using either zero-shot or few-shot operations and using either predicted or oracle belief states.

\subsection{Datasets}
We experiment with two of the currently most prominent benchmark datasets for task-oriented multi-domain dialogue:
\begin{itemize}
    \item \textbf{MultiWOZ 2.2} \cite{budzianowski_multiwoz_2018,hung-etal-2022-multi2woz} is a well-known benchmark used for evaluating state tracking, response generation and dialogue success rate.
    Its evaluation is well-defined and the dataset contains database files, so full interaction can be simulated.
    It contains over 10k dialogues, 7 domains and 29 distinct slots.
    \item \textbf{Schema Guided Dataset} \cite{rastogi2020towards} is also well annotated and even richer dataset containing more than 22k dialogues 18 domains and 145 slots.
    Database interaction is considered in the dataset, but no real database is provided and database results are defined ad-hoc. Therefore we simply use the provided database results in the prompts without performing any actual queries.
\end{itemize}

\subsection{Tested Models}
We chose the following five instruction-finetuned models for our experiments, spanning different sizes (within the limitations of hardware available to us) and using freely available models as well as the paid ChatGPT API.
We indicate the specific model variant (i.e., model size, given by the number of parameters) directly in the model name.
\label{sec:par:models}
\begin{itemize}
    \item \textbf{Tk-Instruct-11B} \cite{supernaturalinstructions} is based on the T5 encoder-decoder architecture \cite{2020t5}. It was tuned on a dataset of over 5M task instances with instructions.
    \item \textbf{ChatGPT} is a product introduced by OpenAI.\footnote{\url{https://openai.com/blog/chatgpt}} Although the exact training process and architectures were not published, it most probably uses a similar architecture and finetuning techniques as InstructGPT \cite{ouyang2022training}, with additional human feedback.
    \item \textbf{Alpaca-LoRA-7B} is a version of the LLaMa model \cite{touvron2023llama} using the LoRA method \cite{hu2021lora} for finetuning on Stanford Alpaca project data \cite{alpaca}. LoRa keeps the base model parameters frozen, but adds additional smaller weight matrices to the model to transform its outputs.
    \item \textbf{GPT-NeoXT-Chat-Base-20B} is based on the GPT-NeoX open-source language model \cite{black2022gpt} and finetuned with over 40M dialogue-style instructions.
    \item \textbf{OPT-IML-30B} \cite{iyer2022opt} is based on the Transformer decoder OPT model \cite{zhang2022opt} and trained with a custom set of instructions, including the finetuning set from Tk-Instruct.
\end{itemize}

\subsection{Evaluated variants}
We test four variants of our setup for each pair of model and dataset.
Specifically, we use zero-shot (without examples) or few-shot (including examples) prompts (\emph{-zs-} vs. \emph{-fs-}) and either generated or oracle belief states (\emph{-gbs} vs. \emph{-obs}).
For retrieval in the few-shot setting, we store just 10 examples per domain in the context store by default. We experiment with increasing this number in Section~\ref{sec:dialogue-performance}.
Using oracle belief state allows us to focus on evaluating the LLM's ability to guide the dialogue.

\begin{figure}[t]
    \centering
    \includegraphics[width=0.49\textwidth]{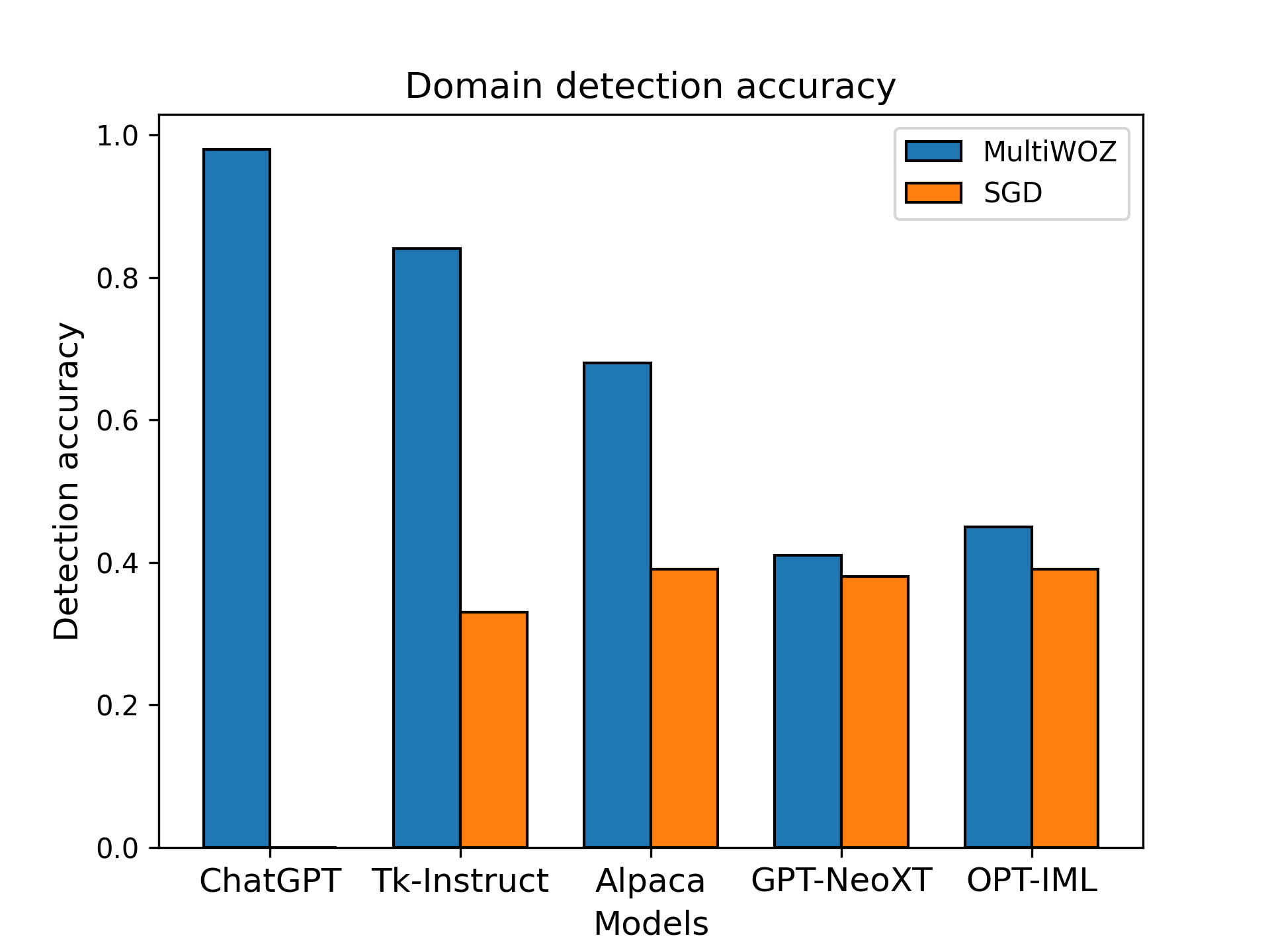}
    \caption{Domain detection accuracy with respect to different models for MultiWOZ 2.2 and SGD data wchich consist of 7 and 18 domains, respectively.}
    \label{fig:domains}
\end{figure}

\subsection{Experiment Details}
\label{subsec:exp-details}
Due to the expensiveness of the LLM runs,\footnote{Hardware intensity for the freely available models and actual cost for ChatGPT.} we did not perform a grid search, but used a limited set of preliminary experiments to determine hyperparameters.
Based on this, we used the context of two preceding  utterances (user + system) as the context store keys (cf.~Section~\ref{sec:context-store}).
We retrieve two examples for few-shot prompts and make one corrupted variant from each of them for negative examples.
To corrupt an example, we switch some of the slot values randomly, similarly to \citet{kulhanek-etal-2021-augpt}.
In the context store, we encode few-shot examples using the multilingual embedding model provided by \citet{reimers-2020-multilingual-sentence-bert}\footnote{\url{https://huggingface.co/sentence-transformers/all-mpnet-base-v2}} and store them in the FAISS database \cite{johnson2019billion}.
To perform the LLM calls, we use the Huggingface library\footnote{\url{https://huggingface.co}} and the OpenAI API.\footnote{\url{https://platform.openai.com}}

\subsection{Evaluation Measures}

We evaluate the system outputs on multiple levels, both using automatic metrics and human evaluation. Results are given in Sections~\ref{sec:results} and~\ref{sec:analysis}, respectively.

\subsubsection*{Automatic Metrics}

In automatic evaluation, we first follow the LLM calls being made and evaluate domain detection, state tracking as well as response generation. We also evaluate the overall dialogue-level performance.
For \emph{domain detection}, we simply compute \textbf{detection accuracy} as a ratio of correctly detected domain out of all dialogue turns being processed.
For \emph{state tracking}, we compute \textbf{micro-F1} score and \textbf{Joint Goal Accuracy} (JGA).
JGA is computed as the ratio of dialogue turns for which the predicted belief state matches the ground truth.
We use fuzzy matching of the slot values, so that capitalization or minor typos do not influence the result.
To evaluate \emph{response generation}, we follow related works and use \textbf{BLEU score} \cite{papineni-etal-2002-bleu}.

The main \emph{overall measure} for evaluating a task-oriented dialogue is the dialogue \textbf{success rate} \cite{deriu_survey_2021}.
For MultiWOZ, we use the standard evaluation of dialogue success as the ratio of dialogues where the user reaches the desired goal, based on goal annotation provided with the data \cite{nekvinda-dusek-2021-shades}. 
The SGD dataset does not include goal annotation but contains information about the requested slots. Therefore, we compute SGD success rate as the proportion of dialogues in which (1) the system captures all the slots correctly and (2) all the requested slots are provided.

\subsubsection*{Human Evaluation}
For human evaluation, we perform a small-scale in-house interaction study on MultiWOZ.
Since the MultiWOZ goal often involves tasks in multiple domains, we ask annotators to evaluate each domain in the dialogue distinctly.
At the end of each dialogue, the annotators are asked to answer these questions:
\begin{enumerate}
    \item \emph{How many of the subdialogues/domains were handled successfully?} (corresponding to dialogue success)
    \item \emph{How many clarifications or corrections were needed?}
    \item \emph{Was all the provided information captured correctly?} (corresponding to JGA)
\end{enumerate}
\section{Automatic Metrics Results}
\label{sec:results}

\subsection{Domain detection}
\label{subsec:domain}
We report the domain detection accuracy on MultiWOZ and SGD
in Figure~\ref{fig:domains}.
We observe that the domain detection accuracy varies quite a lot for various models and presumably influences the quality of the retrieved few-shot examples and appropriateness of the subsequent prompts.
However, it is important to note that domain detection is turn-based, and arguably there are situations (e.g. providing an address, saying goodbye etc.) that are always handled in the same fashion, even though they formally belong to different domains.
Therefore, not all the retrieved examples from misclassified domains necessarily contain unrelated contexts.
To explore this, we measure the performance of all models in case an oracle domain is given to them (Figure \ref{fig:oracle_domains}).
Interestingly, using the oracle domain did not improve performance, it even worsened in some cases.
This suggests that the model-predicted domain is generally good enough, and additionally providing the domain information does not contribute to the final system performance.
The negative influence on performance might be caused by forcing the system to filter out relevant examples.
We observe that in multiple cases, the conversations snippets are domain-independent so the retrieval might perform better even with a wrongly selected domain.
Forcing the ground truth domain examples in these cases can be potentially harmful.

\begin{figure}[t]
    \centering
    \includegraphics[width=0.49\textwidth]{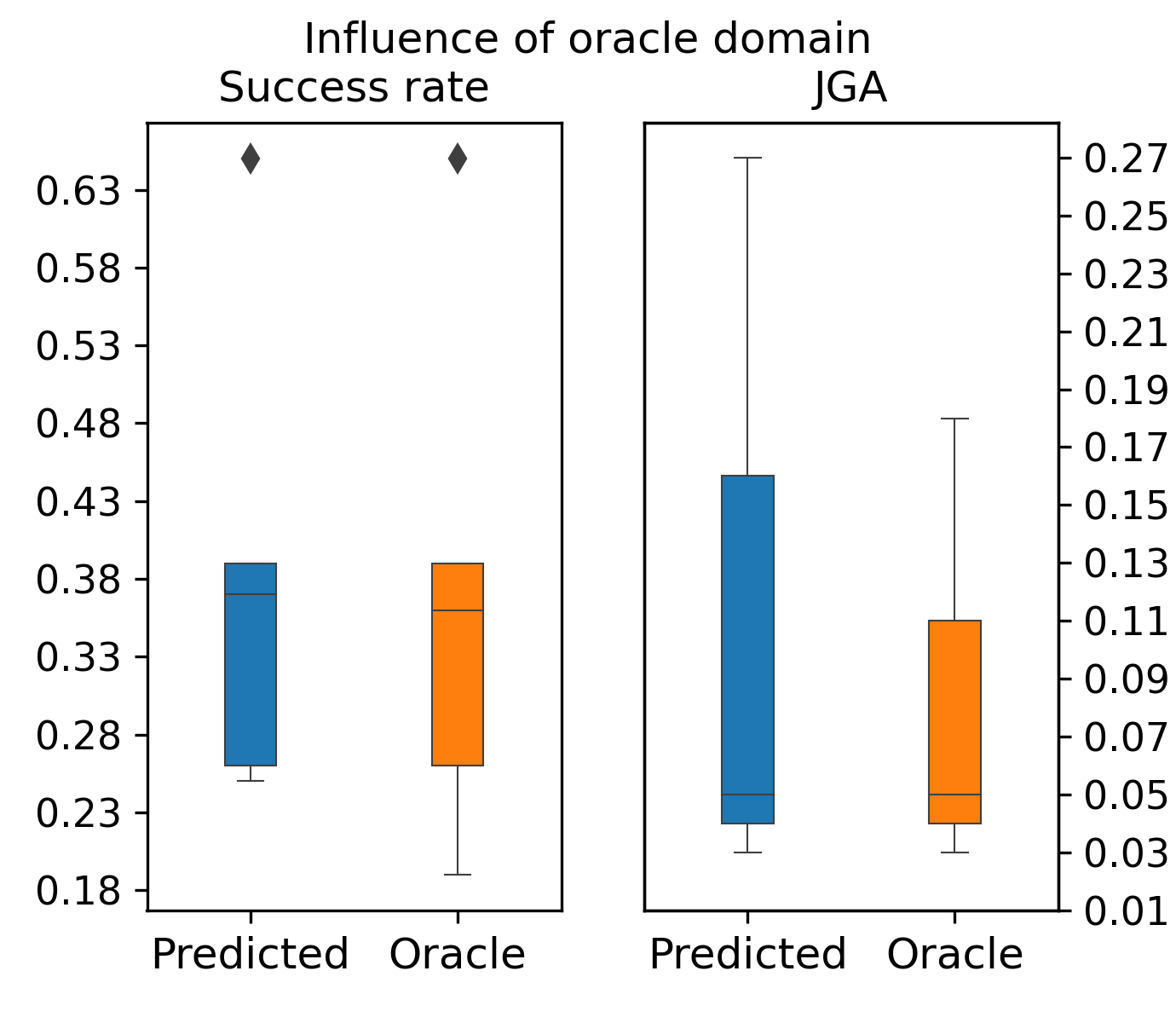}
    \caption{The influence of using oracle domain to retrieve examples. Interestingly, the oracle domain does not improve the performance, suggesting that the model-based detection is good enough for retrieval.}
    \label{fig:oracle_domains}
\end{figure}

\subsection{Belief State Tracking}
\label{subsec:dst}
The belief state tracking results overview is given in Table \ref{tab:res_overall} (\emph{JGA} and \emph{Slot-F1}).
There is a huge gap between the supervised models' performance and the LLM results.
Also compared to \citet{hu-etal-2022-context}, who used few-shot in-context learning and reported JGA 43.13\% with a comparable dataset size, our instruction-tuned LLMs fall short.
However, the models we use are an order of magnitude smaller in general, and we also use fewer examples in the prompt.
We hypothesize that the performance could be further improved by careful model-specific prompt customization and perhaps task re-formulation; nevertheless, this is not the goal of this work.
We intentionally focus on the universal framing of the task since we want to explore the general ability of the models to follow instructions.

When comparing the results among the models, ChatGPT clearly outperforms the rest of the models by a large margin. 
Interestingly, the few-shot vs.\ zero-shot setting does not seem to influence the results much, except for the GPT-NeoXT model.

\subsection{Response Generation}

BLEU scores are low overall, far below the supervised state-of-the-art. Tk-Instruct and ChatGPT are the strongest here and perform roughly on par.

\begin{figure}[t]
    \centering
    \includegraphics[width=0.49\textwidth]{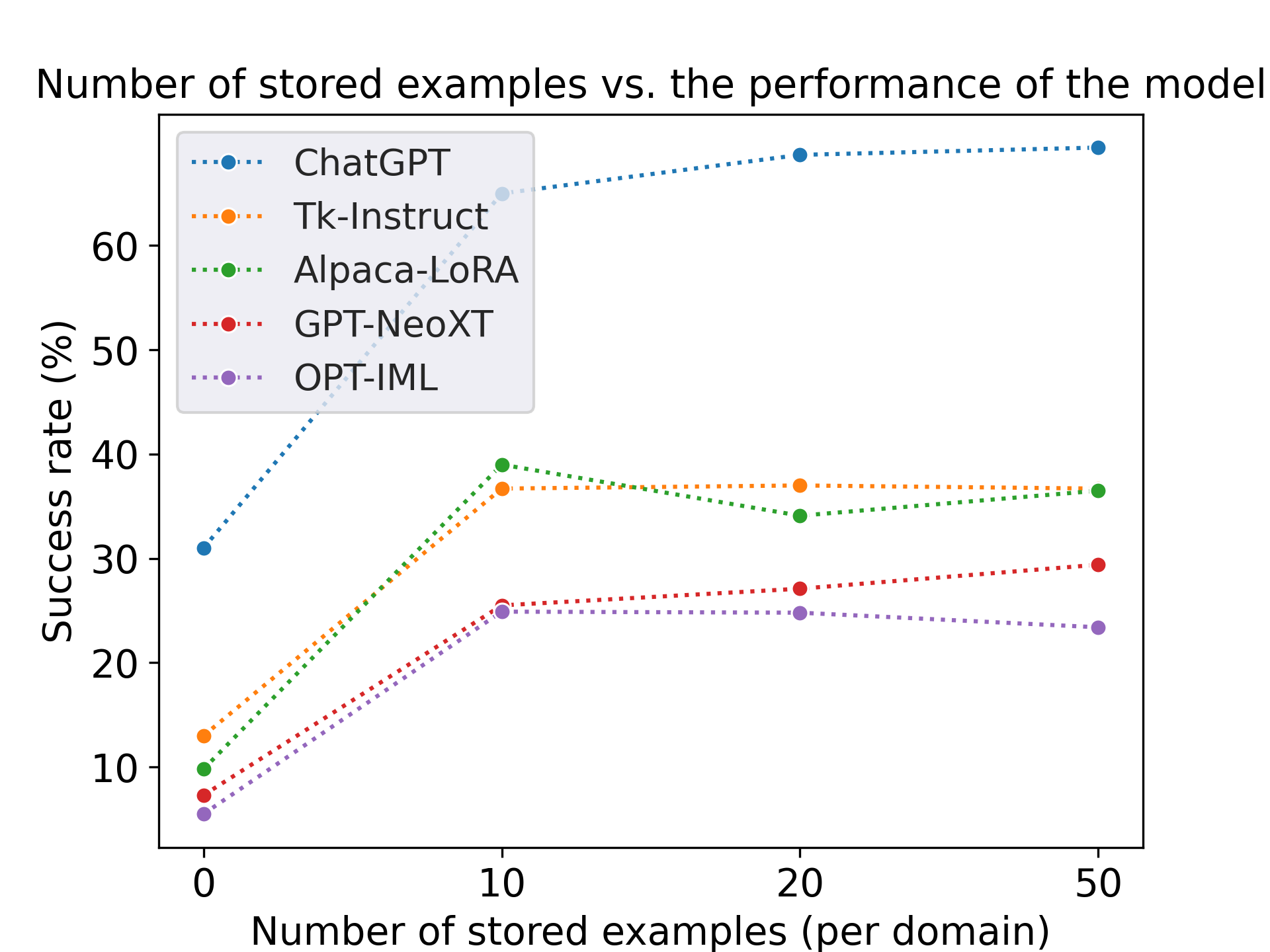}
    \caption{The influence of the number of examples per domain available for few-shot retrieval and performance of the model in terms of the dialogue success on MultiWOZ 2.2 data with oracle state supplied. Note that this does not represent the number of examples selected for the prompt, which is fixed to two.}
    \label{fig:shots}
\end{figure}

\subsection{Dialogue-level performance}
\label{sec:dialogue-performance}

Results for dialogue success are provided in Table~\ref{tab:res_overall}, and there is again a large gap between LLMs and supervised custom models' performance.
ChatGPT seems to outperform other models, similarly to state tracking (cf.~Section~\ref{subsec:dst}).
However, for some cases, especially in the zero-shot setting, the difference is not that obvious.
In most cases, adding the retrieved few-shot examples helps.
The contribution of retrieved examples is more obvious when we supply the oracle belief state, in which case it helps consistently for all the models.

We also explore the influence of context storage size on the dialogue success rate.
The results are given in Figure~\ref{fig:shots}.
It seems that the biggest improvement can be achieved by supplying just a few examples instead of zero-shot prompting, but increasing the size of the example pool for retrieval does not yield further performance gains.

\section{Model Analysis}
\label{sec:analysis}


\subsection{Human Evaluation}

We employed 6 annotators with a background in linguistics and NLP and let them interact with the two strongest models in terms of automatic metrics: ChatGPT and Tk-Instruct.
The annotators were given randomly selected goals from the MultiWOZ 2.2 dataset and a minimal set of essential instructions on how to proceed.
We present the results in Table~\ref{tab:human}.
We can see that in real interaction with a human user and allowing for clarification or correction, the models perform better compared to the rather strict automatic evaluation.
Furthermore, the models are often successful in multiple sub dialogues, even if a part of the whole dialogue fails.
The experiment also confirms the superior performance of ChatGPT on both dialogue success and JGA.
Not surprisingly given the above results, conversations with ChatGPT also required fewer clarification turns than with Tk-Instruct.

\subsection{Error Analysis}

To understand the models' behavior better, we manually inspect a random sample of ca.~20 dialogues for each model, chosen from cases where the automatic success metric was not satisfied. 
In general, we can split most of the erroneous behaviors into two distinct groups, which we call \emph{prompt-recoverable} and \emph{inherent}.

\paragraph{Prompt-recoverable errors} can be likely fixed by specific prompt engineering with some effort.
These errors happen with all of the tested models.
Examples of such errors are the invalid structure of the generated dialogue state, copying slot values instead of using canonical values from the ontology, failure to delexicalize some of the values, etc.
Most of these errors can be also fixed in postprocessing -- for example, we can employ more robust parsers or fuzzy matching of slot values.

\paragraph{Inherent errors,} on the other hand, are likely not easily fixable by prompt modifications.
They are not distributed evenly across the tested models and seem to constitute a more challenging problem.

Perhaps the most important error, common to all the models, is hallucination, i.e., the model's output responses not grounded in the context (such as offering entities that are not included in the database). This happens in about 10-20\% of the inspected dialogues.
Some models (\emph{GPT-NeoXT, OPT-IML}) tend to generate more content than they are asked for.
This happens in more than 50\% of their failed dialogues.
In some cases, this means continuing the conversation for a few more turns (including hallucinating user turns), but the models also often generate unrelated text or even code snippets.
With \emph{Tk-Instruct}, we observed that in ca.~10\% cases, it copies the belief state from the example given in the prompt instead of generating a relevant one.
Another issue is that the models tend to repeat their previous responses.

\begin{table}[t]
    \centering\small
    \begin{tabular}{l|r|r}
    \toprule
    & \textbf{ChatGPT} & \textbf{Tk-Instruct} \\
    \midrule
    dialogues & 25 & 25 \\
    subdialogues & 52 & 48 \\
    clarify / dial & 1.08 & 1.68 \\
    succesful subdialogues & 81\% & 71\% \\
    succesful dialogues & 76\% & 64\% \\
    correctly captured & 88\% & 66\% \\
    \bottomrule

    \end{tabular}
    \caption{Human evaluation results for ChatGPT and Tk-Instruct-11B models. We evaluate the conversation on sub dialogue level i.e. each domain in the dialogue is evaluated separately. }
    \label{tab:human}
\end{table}
\section{Conclusion \& Future Work}
\label{sec:conclusion}

We present an experimental evaluation of instruct\-ion-tuned LLMs applied to the established task of task-oriented dialogue modeling, with five LLMs evaluated on two datasets.
We find that LLMs are not performing well in terms of belief state tracking, even when provided with in-context few-shot examples.
However, there is some potential to improve through prompt tuning and output parsing robust to irregularities.

If provided with a correct belief state, the models can interact with the user successfully, provide useful information and fulfill the user's needs.
While the performance does not match the supervised state of the art, it is important to note that these models were not finetuned on in-domain data and work with just a domain description or a few examples (which again improve performance). 

Therefore, carefully picking representative examples and combining the LLM with an in-domain belief tracker can be a viable choice for a task-oriented dialogue pipeline.

Interestingly, in the human interactive evaluation, both ChatGPT and Tk-Instruct outperformed the expectations set by automatic metrics.
This shows certain flexibility and ability to correct their own mistakes on the part of LLMs, and further demonstrates that single-turn evaluation is too rigid and does not show the whole picture \cite{takanobu_is_2020}.
In future work, we want to focus on addressing the prompt-recoverable errors while maintaining the ability to use model-independent prompts and easily swap models.
We also aim to find a more effective method of relevant example selection.

\section{Limitations}
\label{sec:limit}

One of the limitations of our work is the usage of the ChatGPT model, which is only accessible via an API and is not guaranteed to retain its exact abilities.
However, the other four out of five evaluated models have publicly available weights and their results are fully reproducible.
We still consider it beneficial to also evaluate ChatGPT, as it represents state-of-the-art at the time of writing of this paper and therefore puts the other models' results into perspective.

Another limitation is that based on our empirical experiments, the models are sensitive to the choice of a specific prompting.
We spent some time finding a reasonably good prompt that would work with all of the models and did model-specific modifications for the evaluation.
Specifically, the desired format of the belief state varied between the models, and there we re some model-specific instructions.
We also include both few-shot and zero-shot prompt types in our experiments.
However, it is likely that the performance could be further improved with more extensive prompt engineering efforts.
Nevertheless, we mainly aim to showcase the more raw/out-of-the-box capabilities of the LLMs, as extensive prompt tuning would, in practice, erase the advantage of not having to finetune the models. Furthermore, we believe that the robustness of the model to specific prompts also counts as an added value.

Finally, we cannot exclude the possibility that some of the models were exposed to our selected datasets during training. However, we still find it important to evaluate the LLMs in this setting.

\section*{Acknowledgments}

This work was supported by the project TL05000236 \emph{AI asistent pro žáky a učitele} co-financed by the Technological Agency of the Czech Republic within the ÉTA~5 Programme, 
by the European Research Council (Grant agreement No.~101039303 NG-NLG),
and by the Charles University project SVV 260575.
It used resources provided by the LINDAT/CLARIAH-CZ Research Infrastructure (Czech Ministry of Education, Youth and Sports project No.~LM2018101).
\bibliography{anthology,custom}
\bibliographystyle{acl_natbib}

\clearpage
\onecolumn
\appendix

\section{Prompt Construction}
\begin{table}[H]
    \centering\small
    \begin{tabular}{rl}
      \toprule
      \textbf{Prompt} & {\color{cyan!80!yellow!80!black!100 } Determine which domain is considered in the following dialogue situation. }\\
      &  {\color{green!100!yellow!70!black!100 } Choose exactly one domain from this list: restaurant, hotel, attraction, taxi, train } \\
      & {\color{cyan!80!yellow!80!black!100 } Answer with only one word, the selected domain from the list. You have to always select the most probable domain.} \\
& {\color{red!50!yellow!90!black!100!}  ------- Example 1: -------- } \\
& {\color{red!50!yellow!90!black!100!} Customer: I need a cheap place to eat} \\
& {\color{red!50!yellow!90!black!100!} Assistant: We have several not expensive places available. What food are you interested in?} \\
& {\color{red!50!yellow!90!black!100!} Customer: Chinese food.} \\
& {\color{red!50!yellow!90!black!100!}Domain: restaurant} \\
& {\color{red!50!yellow!90!black!100!} ------ Example 2: -------- } \\
& {\color{red!50!yellow!90!black!100!} Customer: What is the address? } \\
& {\color{red!50!yellow!90!black!100!} Assistant: It's 123 Northfolk Road. } \\
& {\color{red!50!yellow!90!black!100!} Customer: That's all. I also need a train from London. } \\
& {\color{red!50!yellow!90!black!100!} Domain: train } \\
& {\color{red!50!yellow!90!black!100!} ----------- } \\
& {\color{cyan!80!yellow!80!black!100 } Now complete the following example:} \\
& {\color{orange!50!yellow!90!black!100!} Customer: I am looking for a cheap place to stay. }\\
& Domain: \\
      \midrule
      \textbf{Output:} & hotel \\
      \bottomrule
  \end{tabular}
    \caption{A prompt used for domain detection for MultiWOZ.
  It contains {\color{cyan!80!yellow!80!black!100} task definition},  {\color{green!100!yellow!70!black!100!}domains description}, {\color{red!50!yellow!90!black!100!} static examples} and {\color{orange!50!yellow!90!black!100!} user utterance}.}
  \label{app:domain}
\end{table}

\begin{table}[H]
    \centering\small
    \begin{tabular}{rl}
      \toprule
      \textbf{Prompt} & {\color{cyan!80!yellow!80!black!100 }Definition: Capture entity values from last utterance of the conversation according to examples.} \\
    & {\color{cyan!80!yellow!80!black!100 } Capture pair "entity:value" separated by colon and no spaces in between.
    Separate entity:value pairs by hyphens.} \\
      & {\color{cyan!80!yellow!80!black!100!} If not specified, leave the value empty. Values that should be captured are: } \\
      & {\color{green!100!yellow!70!black!100!} - "pricerange": the price of the hotel} \\
      & {\color{green!100!yellow!70!black!100!} - "area" that specifies the area where the hotel is located (north/east/west/south/centre)} \\
      & {\color{green!100!yellow!70!black!100!} - "internet" that specifies if the hotel has internet (yes/no)} \\
      & {\color{green!100!yellow!70!black!100!} - "parking" that specifies if the hotel has parking (yes/no)} \\
      & {\color{green!100!yellow!70!black!100!}- "stars" that specifies the number of stars the hotel has (1/2/3/4/5)} \\
      & {\color{green!100!yellow!70!black!100!} - "type" that specifies the type of the hotel (hotel/bed and breakfast/guest house)} \\
      & {\color{red!100!yellow!70!black!100!}[history] } \\
      &  {\color{orange!50!yellow!90!black!100!}Customer: "I want a cheap place to stay." }\\
      \midrule
      \textbf{Output:} & pricerange:"cheap"\\
      \bottomrule
  \end{tabular}
  \caption{A zero-shot version of the prompt used for state update prediction for MultiWOZ 2.2.
  It contains {\color{cyan!80!yellow!80!black!100} task definition},  {\color{green!100!yellow!70!black!100!}domain description}, {\color{red!100!yellow!70!black!100!} dialogue history} and {\color{orange!50!yellow!90!black!100!} user utterance}. }
  \label{app:zero-shot-state}
\end{table}

\begin{table}[H]
    \centering\small
    \begin{tabular}{rl}
      \toprule
      \textbf{Prompt} & {\color{cyan!80!yellow!80!black!100 }Definition: You are an assistant that helps people to book a hotel.} \\
& {\color{green!100!yellow!70!black!100 }The user can ask for a hotel by name, area, parking, internet availability, or price.} \\
& {\color{green!100!yellow!70!black!100 } There is also a number of hotel in the database currently corresponding to the user's request. }\\
& {\color{green!100!yellow!70!black!100 } If you find a hotel, provide [hotel\_name], [hotel\_address], [hotel\_phone] or [hotel\_postcode]} \\
& {\color{cyan!80!yellow!80!black!100 }Do not provide real entities in the response! Just provide entity name in brackets, like [name] or [address].} \\
& {\color{cyan!80!yellow!80!black!100 } If booking, provide [reference] in the answer. } \\
& {\color{red!100!yellow!70!black!100!}[history] } \\
&  {\color{orange!50!yellow!90!black!100!}Customer: "I want a cheap place to stay." }\\
& {\color{magenta!100!yellow!70!black!100!}State: hotel \{ pricerange: "cheap"\} } \\
& {\color{magenta!100!yellow!70!black!100!} Database: hotels: 23 }\\
      \midrule
      \textbf{Output:} & We have 23 such hotels available, do you have a preference about the location? \\
      \bottomrule
  \end{tabular}
  \caption{A zero-shot version of the prompt used for response prediction for MultiWOZ 2.2.
  It contains {\color{cyan!80!yellow!80!black!100} task definition},  {\color{green!100!yellow!70!black!100!}domain description}, {\color{red!100!yellow!70!black!100!} dialogue history}, {\color{orange!50!yellow!90!black!100!} user utterance} and {\color{magenta!100!yellow!70!black!100!} belief state with db results}.}
  \label{app:zero-shot-response}
\end{table}

\end{document}